\documentclass[runningheads]{llncs}
\usepackage{graphicx}
\usepackage{amsmath,amssymb} %
\usepackage{color}
\usepackage{times}
\usepackage{epsfig}
\usepackage{dsfont}
\usepackage{siunitx}
\usepackage{subfigure}
\usepackage{multirow}
\usepackage{caption}
\usepackage{booktabs}       %
\usepackage{standalone}
\usepackage{xspace}
\usepackage{comment}
\usepackage{hyperref}
\usepackage{harpoon}

\newcommand{\Loss}{\mathcal{L}}

\newcommand{\myparagraph}[1]{\vspace{4pt}\noindent{\bf #1}}

\begin{document}
\title{Grounding Visual Explanations} 

\titlerunning{Grounding Visual Explanations}
\author{Lisa Anne Hendricks\inst{1} \and
Ronghang Hu\inst{1} \and
Trevor Darrell\inst{1}\and
Zeynep Akata\inst{2}}
\authorrunning{L.A. Hendricks, R. Hu, T. Darrell, Z. Akata}

\institute{$^1$ UC Berkeley, $^2$ University of Amsterdam \\
\email{\{lisa\_anne,ronghang,trevor\}@eecs.berkeley.edu} and \email{z.akata@uva.nl}\\
}
\maketitle              %

\begin{abstract}
Existing visual explanation generating agents learn to fluently justify a class prediction.
However, they may mention visual attributes which reflect a strong class prior, although the evidence may not actually be in the image.
This is particularly concerning as ultimately such agents fail in building trust with human users.
To overcome this limitation, we propose a phrase-critic model to refine generated candidate explanations augmented with flipped phrases which we use as negative examples while training. 
At inference time, our phrase-critic model takes an image and a candidate explanation as input and outputs a score indicating how well the candidate explanation is grounded in the image.  
Our explainable AI agent is capable of providing counter arguments for an alternative prediction, i.e. counterfactuals, along with explanations that justify the correct classification decisions.
Our model improves the textual explanation quality of fine-grained classification decisions on the CUB dataset by mentioning phrases that are grounded in the image.
Moreover, on the FOIL tasks, our agent detects when there is a mistake in the sentence, grounds the incorrect phrase and corrects it significantly better than other models. %
\end{abstract}

\section{Introduction}
\label{sec:intro}
Modern neural networks are good at localizing objects~\cite{girshick2015fast}, predicting object categories~\cite{he2016deep} and describing scenes with natural language~\cite{xu2015show}. 
However, the reasoning behind the decision of neural networks are often hidden from the user.  
Therefore, in order to interpret and monitor neural networks, providing explanations of network decisions has gained interest~\cite{hendricks16eccv,guestrin,letham2015interpretable}. 

\begin{figure}[t]
\centering
\includegraphics[width=\linewidth]{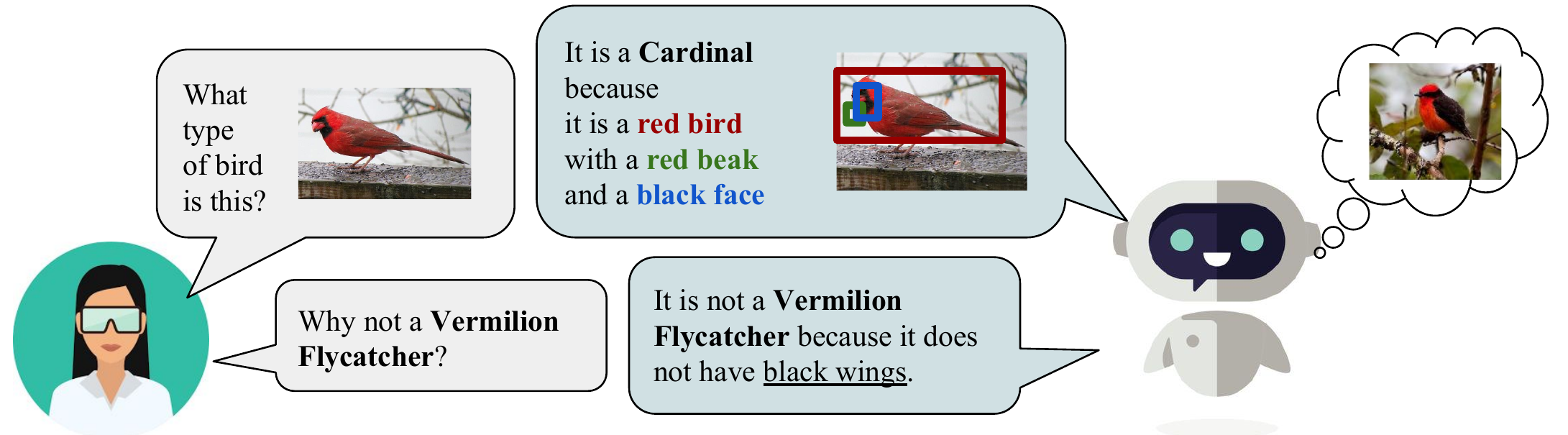}
\caption{Our phrase-critic agent considers grounded visual evidence to determine if candidate explanations are image relevant. 
In this example, as many cardinals are red and have a black patch on their faces, mentioning and grounding those properties constitutes an effective factual explanation, i.e. rationalization. Furthermore, in our framework, informing the user of why an image does not belong to another class via the absence of certain attributes constitutes a counterfactual explanation.}
\label{fig:teaser}
\end{figure}

Ideally, an agent that accurately explains a classifier's decision via natural language, as depicted in~\autoref{fig:teaser}, is expected to generate explanations such as ``This is a \emph{Cardinal} because it is a red bird with a red beak and a black face'' where the phrases should be both \textit{class discriminative}, i.e. a red beak is discriminative for cardinals, and \textit{image relevant}, i.e. the image indeed contains a red beak. 
Moreover, an explainable AI agent should be capable of arguing why the image was not classified as another class such as \emph{Vermilion Flycatcher} by mentioning a class-specific property such as ``black wings'' that differentiates a cardinal from a vermilion flycatcher. 
Contrasting two concepts via class-specific attributes provides an additional means of model interpretation.

Often class and image relevance are in opposition.
For example, if one attribute frequently occurs within a class, an agent may learn to justify its prediction by mentioning this attribute without even looking at the image.
We aim to resolve such conflicts through a phrase-critic which explicitly determines if an explanatory sentence is image relevant by visually grounding discriminative object parts mentioned in an explanation.

One way to design such an agent is to use densely labeled data with ground truth part annotations.
However, it can be time consuming to collect densely labeled data for every task. On the other hand, large, diverse datasets such as the Visual Genome~\cite{krishna2016visual} with densely labeled out-of-domain data do exist.
Detecting visual evidence in a sentence via off-the-shelf grounding models can be unreliable, especially when applied to new domains.
Nevertheless, integrating a natural language grounding model~\cite{hu2017modeling} trained on auxiliary data and an LSTM-based explanation model~\cite{hendricks16eccv} via our proposed \textit{phrase-critic} effectively grounds discriminative phrases in generated explanations.

Our phrase-critic integrates a ranking-loss to the explanation model and builds a set of mismatching part-attribute pairs by flipping attributes in the explanation, inspired by a relative attribute paradigm for recognition and retrieval \cite{relativeattributesiccv}. 
By positing that a bird with a black beak can not also be a bird with a red beak, our model learns to score the correct attribute higher than automatically generated mutually-exclusive attributes. 

We quantitatively and qualitatively show that our \textit{phrase-critic} generates image relevant explanations more accurately than a strong baseline of mean-pooled scores from a natural language grounding model.
Furthermore, our framework can easily be extended to other tasks beyond textual explanations. 
We also show that our phrase-critic framework effectively discerns whether a sentence contains a mistake, points out where the mistake is and fixes it, leading to an impressive performance on FOIL tasks~\cite{shekhar2017foil}.

\section{Related Work}\label{sec:related}
In this section, we review recent papers in the context of explanations, mainly focusing on textual and visual explanations and finally we discuss pragmatics oriented language generation papers that are relevant to ours.

\myparagraph{Explainability.}  The importance of explanations for humans has been studied in the field of psychology~\cite{lombrozo2012,lombrozo2006}, showing that humans use explanations as a guide for learning and understanding by building inferences and seeking propositions or judgments that enrich their prior knowledge. Humans usually seek explanations that fill the requested gap depending on prior knowledge and goal in question. Moreover, explanations are typically contrastive. Much of these ideas are built with careful empirical work, i.e. with human subjects on a specific aspect of explanations~\cite{pacer2013evaluating}. 
Since explanations are intended for a human understander, we emphasize the importance of human evaluation in evaluating the relevance of textual explanations to the image as well as looking for the criteria for what makes an explanation good, with the goal of training a ``critic'' that could evaluate explanations automatically.

\myparagraph{Textual and Visual Explanation.} In \cite{teach1981analysis}, trust is regarded as a primary reason to explore explainable intelligent systems.
We argue a system which outputs discriminative features of an object class without being image relevant is likely to lose the trust of users.
Consequently, we seek to explicitly enforce image relevance with our model.
Like \cite{biran2014justification}, we aim to generate \textit{rationalizations} explaining the evidence for a decision as opposed to introspective explanations which aim to explain the intermediate activations of neural networks.

Early textual explanation models are applied to medical images~\cite{shortliffe1975model} and developed as a feedback for teaching programs~\cite{lane2005explainable,van2004explainable,core2006building}.
These systems are mainly template based. 
Recently, \cite{hendricks16eccv} proposed a deep model to generate natural language justifications of a fine-grained object classifier. 
However, it does not ground the relevant object parts in the sentence or the image. 
In \cite{park2016attentive}, although an attention based explanation system is proposed, there are no constraints to ensure the actual presence of the mentioned attributes or entities in the image. 
Consequentially, albeit generating convincing textual explanations, \cite{hendricks16eccv,park2016attentive} do not include a process for networks to correct themselves if their textual explanation is not well-grounded visually.
In contrast, we propose a general process to first check whether explanations are accurately aligned with image input and then improve textually explanations by selecting a better-aligned candidate.

Other work has considered \textit{visual explanations} which visualize which regions of an image are important for a decision\cite{fong2017interpretable,selvaraju2016grad,zeiler2014visualizing,zintgraf2017visualizing}.
Our model produces bounding boxes around regions which correspond to discriminative features, and is thus visual in nature.  
However, in contrast to visual explanation work, our goal is to rank generated explanatory phrases based on how well they are grounded in an image.

\myparagraph{Pragmatics-Oriented Language Generation.}
Our work is also related to the recent work of pragmatics-oriented language generation~\cite{andreas2016reasoning} where a describer produces a set of sentences, then a choice ranker chooses which sentence best fulfills a specific goal, e.g. distinguishing one image from another.
 Similarly, image descriptions are generated to make the target image distinguishable from a similar image in \cite{vedantam2017context}, and referential expressions are generated on objects in a discriminative way such that one can correctly localize the mentioned object from the generated expression in \cite{mao2016generation}. 
 In this work, we generate textual explanation to maximize both class-specificity and image-relevance. Though similar in spirit, part of our novelty lies in how we learn to rank sentences.

\begin{figure*}[t]
\centering
\includegraphics[width=\linewidth]{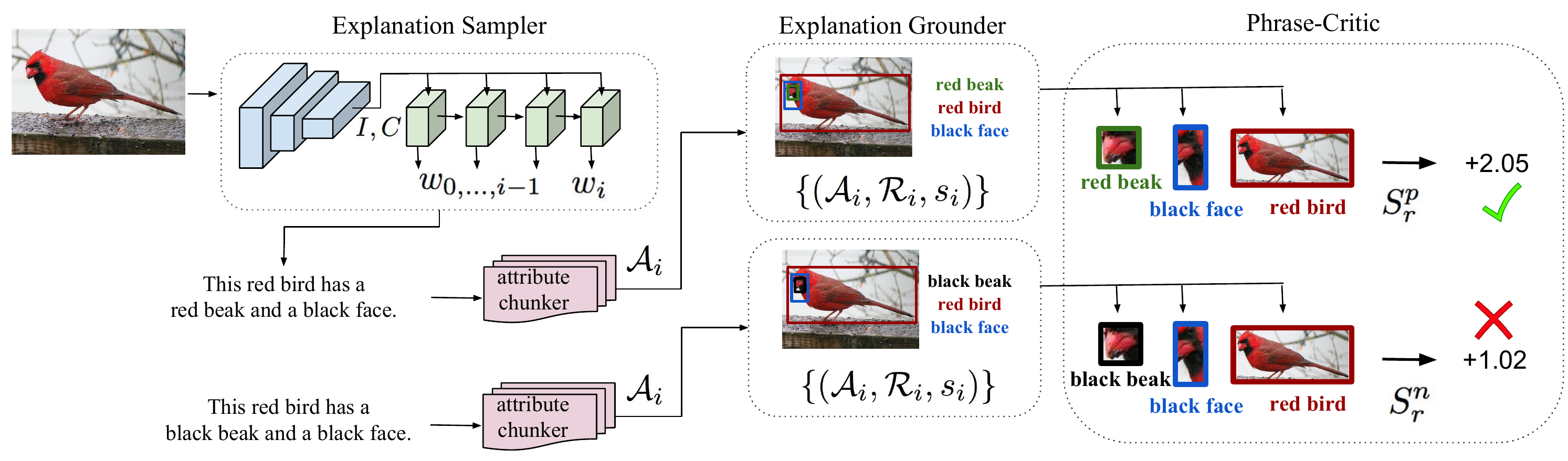}
\caption{Our phrase-critic model ensures that generated explanations are both class discriminative and image relevant. We first sample a set of explanations generated by~\cite{hendricks16eccv}, chunk the sentences into noun phrases and visually ground constituent nouns using~\cite{hu2017modeling}. Our model assigns a score to each noun phrase-bounding box pair and selects the sentence with the highest cumulative score judging it as the most relevant explanation.
}
\label{fig:model}
\end{figure*}

\section{Visual Explanation Critic}\label{sec:model}
Our model consists of three main components.
First, we train a textual explanation model~\cite{hendricks16eccv} with a discriminative loss to encourage sentences to mention class specific attributes.
Next, we train a phrase grounding model~\cite{hu2017modeling} to ground phrases in the generated textual explanations.
Finally, our proposed phrase-critic model ranks textual explanations based on how well they are grounded in the image.
As shown in~\autoref{fig:model}, our system first generates possible explanations (e.g., ``This red bird has a red beak and a black face''), grounds constituent phrases (e.g., ``red bird'', ``red beak'',``black face'') in the image, and then assigns a score to the noun phrase-bounding box pairs based on how well constituent phrases are grounded in the image. The explanation with the highest cumulative score gets selected as the correct explanation.

\myparagraph{Phrase-Critic.} The phrase-critic model constitutes our core innovation. 
Given  a set $\{(\mathcal{A}_i, \mathcal{R}_i, s_i)\}$, where $\mathcal{A}_i$ is an attribute phrase, $\mathcal{R}_i$ is the corresponding region (more precisely, visual features extracted from the region), and $s_i$ the region score, our phrase-critic model, $f_{critic}(\{(\mathcal{A}_i, \mathcal{R}_i, s_i)\})$, maps them into a single image relevance score $S_r$. 
For a given attribute phrase $A_i$ such as ``black beak'', we ground (localize) it into a corresponding image region $R_i$ and obtain its localization score $s_i$, using an off-the-shelf localization model from \cite{hu2017modeling}.  
It is worth noting that the scores directly produced by the grounding model can not be directly combined with other metrics, such as sentence fluency, because these scores are difficult to normalize across different images and different visual parts.
For example, a correctly grounded phrase ``yellow belly'' may have a much smaller score than the correctly grounded phrase ``yellow eye'' because a bird belly is less well defined than a bird eye. 
Henceforth, our phrase-critic model plays an essential role in producing normalized, utilizable and comparable scores. 
More specifically, given an image $I$, the phrase-critic model processes the list of $\{(\mathcal{A}_i, \mathcal{R}_i, s_i)\}$ by first encoding each $(\mathcal{A}_i, \mathcal{R}_i, s_i)$ into a fixed-dimensional vector $x_{enc}$ with an LSTM and then applying a two-layer neural network to regress the final score $S_r$ which reflects the overall image relevance of an explanation.

We construct ten negative explanation sentences for each image as we explain in the next section. 
Each negative explanation sentence (not image relevant) gets paired with a positive explanation (image-relevant). 
We then train our explanation critic using the following margin-based ranking loss $\Loss_\mathrm{rank}$ on each pair of positive and negative explanations, to encourage the model to give higher scores to positive explanations than negative explanations:
\begin{align}
\Loss_\mathrm{rank} %
& = \max(0, \underbrace{f_{critic}(\{A_{i}^{n}\}, I; \theta)}_{S_r^n} - \underbrace{f_{critic}(\{A_{i}^{p}\}, I; \theta)}_{S_r^p} + 1)
\end{align}
where $A_{i}^{p}$ are matching noun phrase whereas $A_{i}^{n}$ are mismatching noun phrases respectively, therefore $S_r^p$ and $S_r^m$ are the scores of the positive and the negative explanations. %
In the following, we discuss how we construct our negative image-sentence pairs.
 
\myparagraph{Mining and Augmenting Negative Sentences.} 
The simplest way to sample a negative pair is to consider a mismatching ground truth image and sentence pair.
However, we find that mismatching sentences are frequently either too different from ground truth sentences (and thus do not provide a useful training signal) or too similar to ground truth sentences, such that both the positive and negative sentence are image relevant. 
Hence, inspired by a relative attribute paradigm for recognition and retrieval \cite{relativeattributesiccv}, we create negative sentences by flipping attributes corresponding to color, size and objects in attribute phrases.  
For example, if a ground truth sentence mentions a ``yellow belly'' and ``red head'' we might change the attribute phrase ``yellow belly'' to ``yellow beak'' and ``red head'' to ``black head''.  
This means the negative sentence still mentions some attributes present in the image, but is not completely correct. 
We find that creating hard negatives is important when training our self-verification model.

\myparagraph{Ranking Explanations.}
After generating a set of candidate explanations and computing an explanation score, we choose the best explanation based on the score for each explanation.
In practice, we find it is important to rank sentences based on both the relevance score $S_r$ and a fluency score $S_f$ (defined as the $\log P(w_{0:T})$). 
However, we find that first discarding sentences which have a low fluency score, and then choosing the sentence with the highest relevance works better:
\begin{equation}
S = \mathds{1}\bigg(\underbrace{\sum_{i} \log P(w_i | w_{0,...,i-1})}_{S_f} > T\bigg)\underbrace{f_{critic}(\{A_i\}, I; \theta)}_{S_r} 
\end{equation}
where $S_r$ is the relevance score and $S_f$ is the log probability of a sentence based on the trained explanation model.  $\mathds{1}(\cdot)$ is the indicator function and $T$ is a fluency threshold.
Including $S_f$ is important because otherwise the explanation scorer will rank ``This is a bird with a long neck, long neck, and red beak'' high (if a long neck and red beak are present) even though mentioning ``long neck'' twice is clearly ungrammatical.
Based on experiments on our validation set, we set $T$ equal to negative five.

\myparagraph{Grounding Visual Features.} Our framework for grounding visual features involves three steps: generating visual explanations, factorizing the sentence into smaller chunks, and localizing each chunk with a grounding model. 
Visual explanations are generated with a recurrent neural network (specifically an LSTM \cite{hochreiter1997long}) over the image.
Unlike standard visual description models, e.g. \cite{donahue16pami}, the visual explanation generation model~\cite{hendricks16eccv} is conditioned on the class $C$ predicted by the visual model as well as the image $I$ itself.
The explanation model relies on two losses: a relevance loss 
\begin{equation}
\Loss_{\mathrm{rel}} = \frac{1}{N} \sum_{n=0}^{N-1}\sum_{t=0}^{T-1} \log p(w_{t+1}|w_{0:t}, I, C)
\end{equation}
which corresponds to the standard word level softmax cross entropy loss used to train sentence generation models, and a discriminative loss
\begin{equation}
\Loss_{\mathrm{discr}} = \mathbb{E}_{\tilde{w} \sim p(w|I,C)} \left[ R(\tilde{w}) \right]
\end{equation}
which assigns a high reward ($R$) to class discriminative features sentences. 
REINFORCE~\cite{williams1992simple} is used to backpropagate through the sampling mechanism required to generate sentences for the discriminative loss.
The visual explanation generation loss ($\Loss_{\mathrm{VEG}}$) is the linear combination of these two losses:
\begin{equation}
\Loss_{\mathrm{VEG}} = \Loss_{\mathrm{rel}}  - \lambda \Loss_{\mathrm{discr}}.
\end{equation}

In order to verify that explanations are image relevant, for each explanation we extract a list of $i$ attribute phrases ($\mathcal{A}_i$) using a rule-based attribute phrase chunker. 
Our chunker works as follows: we first use a POS tagger, then extract attribute phrases by finding phrases which syntactically match the structure of attribute phrases. 
We find that attribute phrases have two basic types of syntactic structure: a noun followed by a verb and an adjective, e.g. ``bird is black'' or ``feathers are speckled'', or an adjective (or list of adjectives) followed by a noun, e.g. ``red and orange head'' or ``colorful body''.
Though this syntactic structure is specific to the bird data, similar methods could be used to extract visual phrases for other applications.
Attribute phrases are ordered based on the order in which they occur in the generated visual explanation.

Once we have extracted attribute phrases $\mathcal{A}_i$, we ground each of them to a visual region $\mathcal{R}_i$ in the original image by using \cite{hu2017modeling} pre-trained on the Visual Genome dataset \cite{krishna2016visual} without any access to task-specific ground truth.
For a given attribute phrase $\mathcal{A}_i$, the grounding model localizes the phrase into an image region, returning a bounding box $\mathcal{R}_i$ and a score $s_i$ of how likely the returned bounding box matches the phrase. 
The grounding model works in a retrieval manner. It first extracts a set of candidate bounding boxes from the image, and embeds the attribute phrase into a vector. Then the embedded phrase vector is compared with the visual features of each candidate bounding box to get a matching score ($s_i$). Finally the bounding box with the highest matching score is returned as the grounded image region.
The attribute phrase, the corresponding region, and the region score form an attribute phrase grounding $(\mathcal{A}_i, \mathcal{R}_i, s_i)$.
This attribute phrase grounding is used as an input to our phrase-critic.

Whereas visual descriptions are encouraged to discuss attributes which are relevant to a specific class, the grounding model is only trained to determine whether a natural language phrase is in an image. 
Being discriminative rather than generative, the critic model does not have to learn to generate fluent, grammatically correct sentences, and can thus focus on checking whether the mentioned attribute phrases are image-relevant. 
Consequently, the models are complementary, allowing one model to catch the mistakes of the other.

\section{Experiments}\label{sec:experiments}
In this section, we first detail the datasets we use in our experiments. 
We then present a comparison between different methods to rank sampled sentences.  
We compare our model to baselines both qualitatively and through a human evaluation.
Additionally, we discuss how our model enables \textit{counterfactual} explanations.

\myparagraph{CUB.} We validate our approach on the CUB dataset~\cite{welinder10tr} which contains 200 classes of fine-grained bird species with approximately 60 images each and a total of 11,788 images of birds. Recently,~\cite{RALS16} collected 10 sentences for each image with a detailed description of the bird. This dataset is suitable for our task as every sentence as well as every image is associated with a single label. Note that CUB does not contain ground truth part bounding boxes, however it contains keypoints that roughly fall on each body part. We use them only to evaluate the precision of our detected bounding boxes.
To generate sentences, we use random sampling~\cite{donahue16pami} to sample 100 sentences from the baseline model proposed by~\cite{hendricks16eccv}.
We use the set of 100 sentences generated via random sampling as candidate sentences for phrase-critic.
As a ``Baseline'' model, we select sentences based only on the fluency score $S_f$.
In addition to our phrase-critic, we also consider a strong baseline which only considers the score from a grounding model (without the phrase critic), i.e. we call it ``Grounding model''.
In this case, we first ground noun phrases, then rank sentences by the average score of the grounded noun phrases in the image.

\myparagraph{FOIL.} Our phrase-critic model is flexible and can also be applied to other relevant tasks.
To show the generality of our approach, we also consider the dataset proposed in \cite{shekhar2017foil} which consists of sentences and corresponding ``FOIL'' sentences which have exactly one error.
\cite{shekhar2017foil} proposes three tasks: (1) classifying whether a sentence is image relevant or not, (2) determining which word in a sentence is not image relevant and (3) correcting the sentence error.
To use our phrase-critic for (1), we employ a standard binary classification loss.
For (2), we follow \cite{shekhar2017foil} and determine which words are not image relevant by holding out one word at a time from the sentence.  When we remove an irrelevant word, the score from the classifier should increase.  
Thus, we can determine the least relevant word in a sentence by observing which word (upon removal) leads to the largest score from our classifier.
Also following \cite{shekhar2017foil}, for the third task we replace the foiled word with words from a set of target words and choose a target word based on which one maximizes the score of the classifier.
To train our phrase critic, we use the positive and negative samples as defined by \cite{shekhar2017foil}.
As is done across all experiments, we extract phrases with our noun phrase chunker and use this as input to the phrase-critic.

\begin{figure*}[t]
\centering
\includegraphics[width=\linewidth]{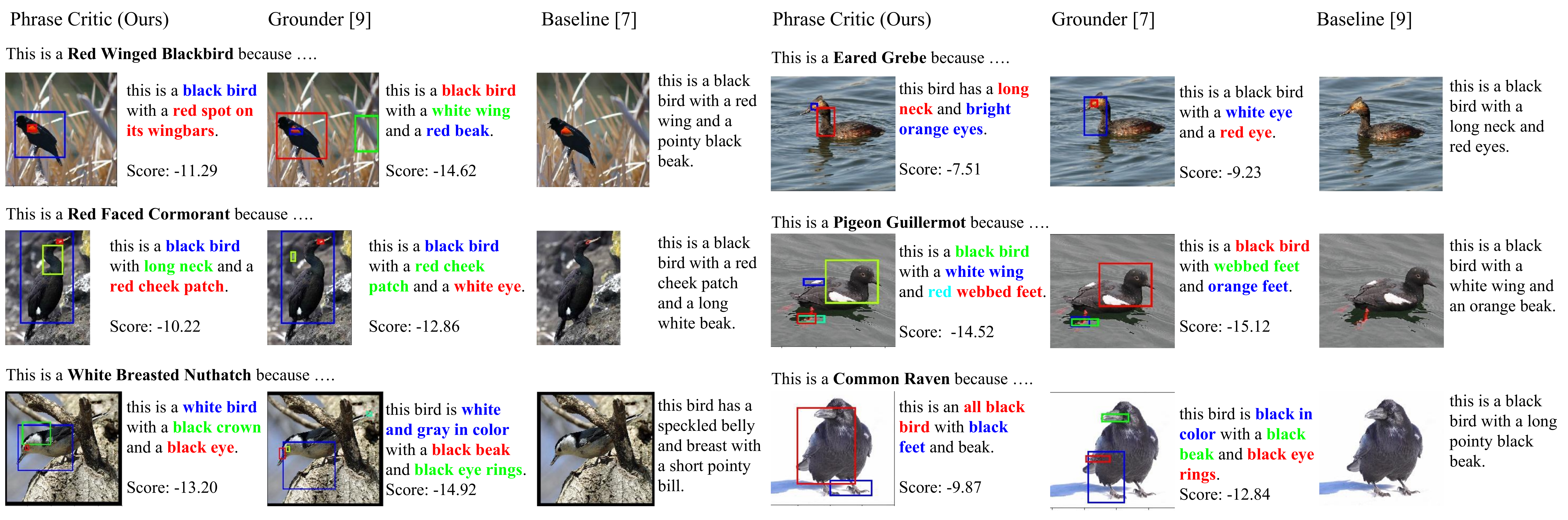}
\caption{Our phrase-critic model generates more image-relevant explanations compared to~\cite{hendricks16eccv} justified by the grounding of the noun phrases. Compared to Grounder~\cite{hu2017modeling}, our phrase-critic generates more class-specific explanations. The numbers indicate the cumulative score of the explanation computed by our phrase-critic ranker.}
\label{fig:CriticVBaseline2}
\end{figure*}

\subsection{Fine-Grained Bird Species Explanation Experiments}

In this section, we conduct detailed bird species explanation experiments on the CUB dataset. We first present comparison with baselines qualitatively both for successful and failure cases as well as quantitatively through human judgement. We then present results illustrating the accuracy of the detected bounding boxes. We finally discuss our counterfactual explanation results.

\myparagraph{Baseline Comparison.}
Our aim here is to compare our phrase-critic model with the baseline visual explanation model~\cite{hendricks16eccv} and the grounding model~\cite{hu2017modeling}. 
In~\autoref{fig:CriticVBaseline2}, the results on the left are generated by our phrase critic model, the ones in the middle by the grounding model~\cite{hu2017modeling} and the ones on the right are by the baseline model~\cite{hendricks16eccv}.
Note that \cite{hendricks16eccv} does not contain an attribute phrase grounder, therefore we cannot localize the evidence for the given explanation here. 
As a general observation, our model improves over both baselines in the following ways. Our critic model (1) grounds attribute phrases both in the image and in the sentence, (2) is in favor of accurate and class-specific noun phrases and (3) provides the cumulative score of each explanatory sentence. 

To further emphasize the importance of visual and textual grounding of the noun phrases in evaluating the accuracy of the visual explanation model, let us more closely examine the second row of~\autoref{fig:CriticVBaseline2}.
We note that all models mention a ``black bird'' and ``red cheek patch''.
As the ``Red Faced Cormorant'' has these properties, these attributes are accurate. 
However, the explanation sentence is more trustable when the visual evidence of the noun phrase properly localized, which is not done by the baseline explanation model.  
To verify our intuition that grounded explanations are more trustable, we ask Amazon Mechanical Turke workers to evaluate whether our explanations with or without bounding boxes are more informative.
Our results indicate that bounding boxes are informative  (41.9\% of the time bounding boxes lead to more informative explanations and 49.3\% of the time explanations with and without bounding boxes are equally informative).
Therefore, we emphasize that visual and textual grounding is beneficial and important for evaluating the accuracy of the visual explanation model. 
Again examining the ``Red Faced Cormorant'' in the second row of~\autoref{fig:CriticVBaseline2}, although ``red cheek patch'' is correctly grounded both by our phrase critic and the baseline phrase grounder, our phrase critic also mentions and grounds an important class-specific attribute of ``long neck'' while the grounding model mentions a missing ``white eye'' attribute which it cannot grouned. 
Thus, the score based ranking of noun phrase and region pairs lead to more accurate and visually grounded visual explanations.

Thanks to the integrated visual grounding capability and phrase ranking mechanism, the critic is able to detect the mistakes of the baseline model and correct them. Some detailed observations from~\autoref{fig:CriticVBaseline2} are as follows. 
``Red Winged Blackbird'' having a ``red spot on its wingbars'' is one of the most discriminative properties of this bird which is mentioned by our critic and also grounded accurately. 
Similarly, the most important property of ``Eared Grebe'' is its ``red eyes''.
We see that for ``Pigeon Guillermot'' our model talks about its ``white wing'' and ``red webbed feet'' whereas the grounding model does not mention the  ``white wing'' property and the baseline model does not only ground the phrase but also it does not mention the ``red feet''. 
Our model does not only qualitatively generate more accurate explanations, these sentences also get higher cumulative phrase scores as shown beside each image in the figure providing another level of confidence.

\begin{figure*}[t]
\centering
\includegraphics[width=\linewidth]{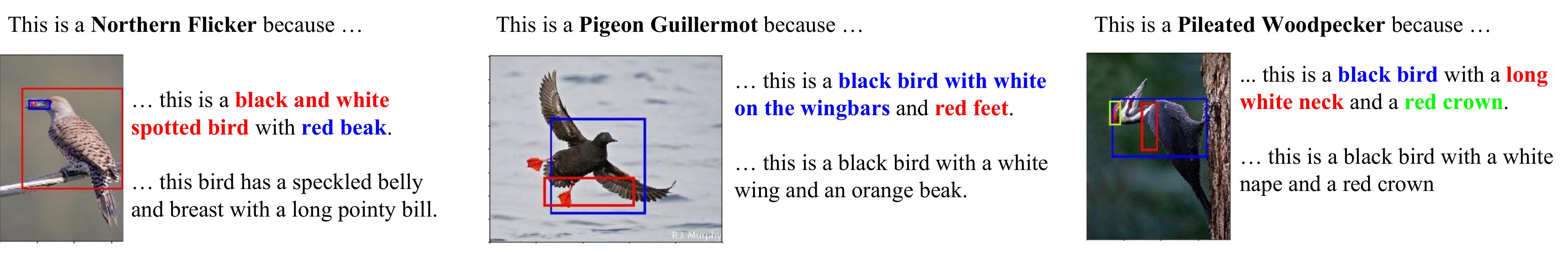}
\caption{Failure Cases: In some cases our model predicts an incorrect noun phrase and the grounding may reveal the reason. On the other hand, in some cases although the explanation sentences are accurate, the phrases are not grounded well, i.e. the bounding boxes are off. Top: Our phrase-critic, Bottom: Baseline~\cite{hendricks16eccv}.}
\label{fig:CriticVBaseline3}
\end{figure*}

\myparagraph{Failure Cases.} In \autoref{fig:CriticVBaseline3} we present some typical failure cases of our model. In some cases such as the first example, the nouns, i.e. bird and beak, are correctly grounded however the attribute is wrong. Although the bird has a black beak, due to the red color of the fruit it is holding, our model thinks it is a red beak. 
Another failure case is when the noun phrases are semantically accurate however they are not correctly grounded. For instance, in the second example, both ``black bird with white on the wingbars'' and ``red feet'' are correctly identified, however the bounding box of the feet is off. Note that in CUB dataset, the ground truth part bounding box annotations are not available, hence our model figures out the location of a ``red feet'' by adapting the grounding model trained on Visual Genome, which may not include similar  box-phrase combinations. Similarly, in the third example, the orientation of the bounding box of the phrase ``long white neck'' is inaccurate since the bird is perching on the tree trunk vertically although most of the birds perch on tree branches in a horizontal manner.

{
\setlength{\tabcolsep}{5pt}
\renewcommand{\arraystretch}{1.2}
\begin{table}[t]
\centering
\begin{tabular}{ l  c   c }
Method & \% Correct Noun Phrases & \% Correct Sentences \\  
\hline
 Baseline~\cite{hendricks16eccv} &  76.64 &   52.10  \\  
 Grounding model~\cite{hu2017modeling} &  76.32 &  49.85   \\    
 Our Phrase Critic &  \textbf{77.96} & \textbf{61.97}     
\end{tabular}
\caption{Human evaluations comparing Baseline explanation model~\cite{hendricks16eccv}, Grounding model~\cite{hu2017modeling}, and our phrase-critic model.  CNP: the percentage of correct noun phrases predicted by each model, CS: the percentage of correct sentences where all the phrases are semantically accurate. } 
\label{tab:humanevaluation}
\end{table}
}

\myparagraph{Human Evaluation.} 
In this section we discuss the effectiveness of our explanation ranker via human evaluation.  
We sample 2000 random images from the test set and ask Amazon Mechanical Turkers to annotate whether a noun phrase selected by a model is observed in the image or not.
We run this human study three times to eliminate the annotator bias. 
In this study, we measure the percentage of correct noun phrases, i.e. CNP, and the correct sentences, i.e. CS, that are agreed by at least the two out of three annotators.
Our results are presented in~\autoref{tab:humanevaluation}.  
We first compare our phrase-critic method to the baseline explanation generator~\cite{hendricks16eccv}.
We find that sentences chosen with our phrase critic have improved CNP score ($77.96\%$ vs $76.64\%$) in contrast to our baseline.
Next, we find that attributes mentioned by our critic model reflect the images more accurately than the grounding only baseline~\cite{hu2017modeling} ($77.96\%$ image relevant attributes vs. $76.32\%$).
This result shows that our phrase critic model generates phrases that are more accurate than the baselines as well as being visible in the image.

We also compute the percentage of correct sentences (CS), i.e. all noun phrases appearing in the sentence correctly match the given image, for each model.
We observe the same ordering trend between the three models as the CNP scores, i.e. Grounding model $<$ Baseline $<$ Phrase Critic with the phrase-critic producing correct sentences 59.67\% of the time in comparison to 51.49\% and 47.74\% for the baseline and grounder respectively.
We note that the performance gain of our phrase-critic model is larger when considering entire sentences, perhaps because our phrase-critic is specifically optimized to discriminate between sentences.
These results indicate that our phrase critic model leads to fewer mistakes overall. 

\myparagraph{Bounding Box Accuracy.} As the CUB dataset does not contain ground-truth bounding boxes, we cannot evaluate the precision of our detected part bounding boxes w.r.t. a ground truth. 
However, the dataset contains keypoints for 15 body parts, e.g. bill, throat, left eye, nape, etc. and utilizing these keypoint annotations that roughly correspond to ``beak'', ``head'', ``belly'' and ``eye'' regions, provides us a good proxy for this task. We measure how frequently a keypoint falls into the detected bounding box of the corresponding body part to determine the accuracy of the bounding boxes. In addition, we measure the distance of the corresponding keypoint to the center of the bounding box to determine the precision of the bounding boxes. Note that for the results in the first row, we take the explanation generated by~\cite{hendricks16eccv} and ground the phrases using the off-the-shelf grounding model~\cite{hu2017modeling}.

{
\setlength{\tabcolsep}{5pt}
\renewcommand{\arraystretch}{1.2}
\begin{table}[t]
\centering
\begin{tabular}{ l   c   c  c  c | c c c c}
& \multicolumn{4}{c}{\% Accuracy} & \multicolumn{4}{c}{Euclidean Distance} \\
Explanations & Beak & Head & Belly & Eye  & Beak & Head & Belly & Eye \\ \hline
Baseline~\cite{hendricks16eccv} & 93.50 & 58.74 & 65.58 & 55.11 & 24.16 & 57.56 & 56.80& 76.90\\
Grounding model~\cite{hu2017modeling} & 94.30 & 60.60 & 65.40 & \textbf{60.78} &\textbf{22.66} & 46.31& \textbf{52.69} & \textbf{57.55} \\
Phrase Critic & \textbf{95.88} & \textbf{74.06} & \textbf{66.65} & 56.72 & 23.74 & \textbf{20.26} & \textbf{52.75} & 69.83 \\
\end{tabular}
\caption{Evaluating the grounding accuracy for four commonly mentioned bird parts. 
As we have no have access to ground truth boxes, we measure how frequently the ground truth keypoints fall within a detected bounding box, measuring the \% of the keypoints that fall inside the bounding box (left) and the distance between the keypoint and the center of the bounding box (right).
The Baseline~\cite{hendricks16eccv} does not include noun phrase grounding, so we apply~\cite{hu2017modeling} to noun phrases extracted from \cite{hendricks16eccv}.
}
\label{tab:keypoint}
\end{table}
}

Our results in~\autoref{tab:keypoint} show that while ``beak'', ``head'' and ``belly'' regions are detected with high accuracy ($95.88\%$, $74.06\%$ and $66.65\%$ resp.), ``eye'' detections are weaker ($56.72\%$).
When we look at the distance between the bounding box center and the keypoint, we observe a similar trend. 
The head region gets detected by our model significantly better than others, i.e. $20.26$ vs $46.31$ with~\cite{hu2017modeling} and $57.56$ with~\cite{hendricks16eccv}. 
The belly and the beak distances are close to the ones measured by the grounding model whereas the eye region gets detected with a lower precision with our model compared to the grounding model.

We closely investigate the accuracy of the predicted noun phrases that fall into the eye region and observe that although the eye regions get detected with a higher precision with the baseline grounding model, the semantic meaning of the attribute gets predicted more accurately with our phrase critic. For instance, our model mentions ``red eye'' more accurately than the grounding model although the part box is more accurately localized by the grounding model.
One example of this can be seen in~\autoref{fig:CriticVBaseline2} (top right) where the grounder selects the sentences ``... this is a black bird with a white eye and a red eye.''
Here, the grounding model has selected a sentence which cannot be true (the bird cannot have white eye as well as a red eye).
Even though the bounding box around the eye is accurate, the modifying attributes are not both correct.

\myparagraph{Counterfactual Explanations.}
Another way of explaining a visual concept is through generating \textit{counterfactual} explanations that indicate why the classifier does not predict another class label.
To construct counterfactual explanations, we posit that if an attribute is discriminative for another class, i.e. a class that is different from the class that the query image belongs to, but not present in the query image, then this attribute is a \textit{counterfactual} evidence. 
To discuss counterfactual evidence for a classification decision, we first hypothesize which visual evidence might indicate that the bird belongs to another class.
We do so by considering explanations produced by our phrase-critic for visually most similar examples from a different, i.e. counterfactual, classes.
Our phrase-critic determines which attributes are most class specific for the counterfactual class and most image relevant for the query image while generating factual explanations. 
While generating counterfactual explanations, our model determines the counterfactual evidence by searching for the attributes of the counterfactual class which lead to the lowest phrase-critic score for the query image. 
We then construct a sentence by negating counterfactual phrases. For instance, ``bird has a long flat bill'' is negated to ``bird does not have a long flat bill'' where the counterfactual phrase is the ``long flat bill''. Alternatively, we can use the same evidence to rephrase the sentence ``If this bird had been a (counterfactual class), it would have had a long flat bill.''

\begin{figure*}[t]
\centering
\includegraphics[width=\linewidth]{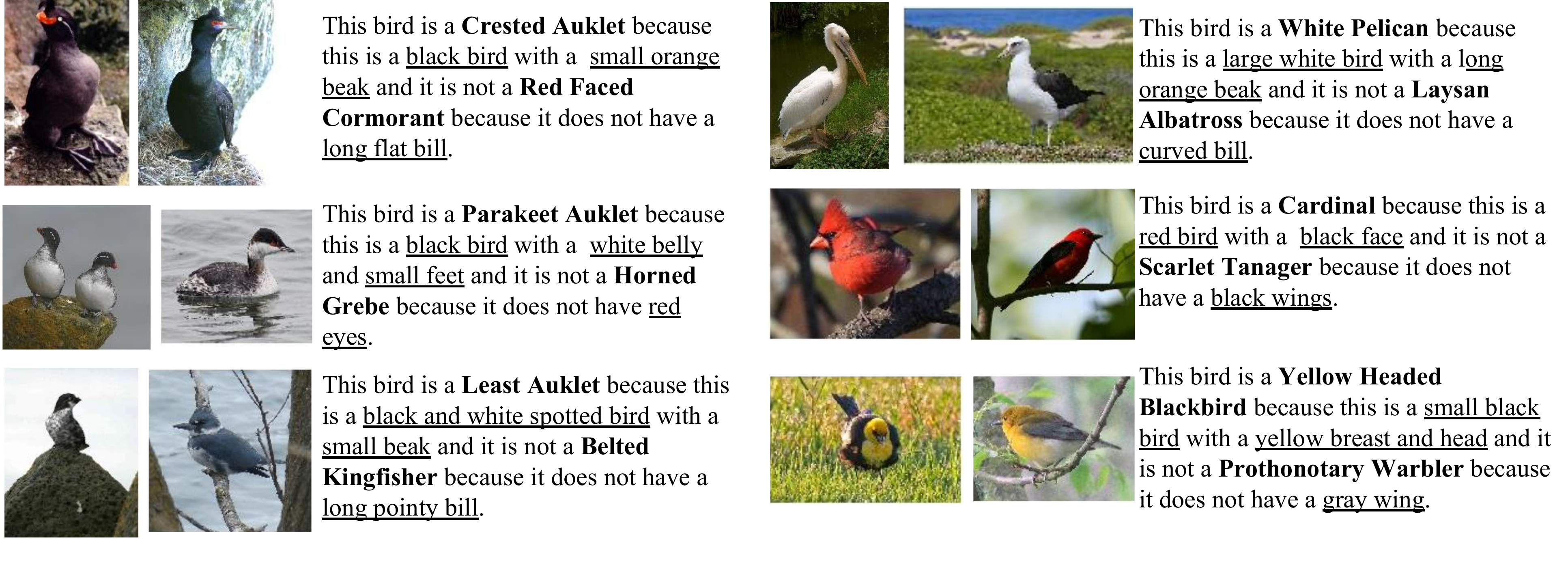}
\caption{Our phrase-critic is able to generate factual and counterfactual explanations. Factual explanations mention the characteristic properties of the correct class (left image) and counterfactual explanations mention the properties that are not visible in the image, i.e. non-groundable properties, for the negative class (right image).}
\label{fig:counterfactual}
\end{figure*}

To illustrate, we present our results in \autoref{fig:counterfactual}. 
Note that the figures show two images for each result where the first image is the query image. 
The second image is the counterfactual image, i.e. the most similar image to the query image from the counterfactual class, that we show only for reference purposes. 
The counterfactual explanation is generated for this image just for determining the most class-specific noun phrase.
Once a list of counterfactual noun phrases is determined, those noun phrases are grounded in the query image and the noun phrase that gets the lowest score is determined as the counterfactual evidence. 
To illustrate, let us consider an image of a \emph{Crested Auklet} and a nearest neighbor image from another class, e.g., \emph{Red Faced Cormorant}.
The attributes ``black bird'' and ``long flat bill'' are possible counterfactual attributes for the original crested auklet image.
We  use our phrase-critic to select the attribute which produces the \textit{lowest} score for the Crested Auklet image.

\autoref{fig:counterfactual} shows our final counterfactual explanation for why the \emph{Crested Auklet} image is not a \emph{Red Faced Cormorant} (it does not have a long flat bill).
On the other hand, when the query image is a \emph{Parakeet Auklet}, the factual explanation talks about ``red eyes'' which are present for \emph{Horned Grebe} but not for \emph{Parakeet Auklet}.  
Similarly, a \emph{Least Auklet} is correctly determined to be a ``black and white spotted bird'' with a ``small beak'' while a \emph{Belted Kingfisher} is a has a ``long pointy bill'' which is the counterfactual attribute for \emph{Least Auklet}. 
On the other hand, a \emph{Cardinal} is classified as a cardinal because of the ``red bird'' and ``black face'' attributes while not as a \emph{Scarlet Tanager} because of the lack of ``black wings''. 
These results show that our counterfactual explanations do not always generate the same phrases for the counterfactual classes. 
Our counterfactual explanations talk about properties of the counterfactual class that are not relevant to the particular query image, whose evidence is clearly visible in both the counterfactual and the query images.

In conclusion, counterfactual explanations go one step further in language-based explanation generation. 
Contrasting a class with another closely related class helps the user build a more coherent cognitive representation of a particular object of interest.

{
\setlength{\tabcolsep}{5pt}
\renewcommand{\arraystretch}{1.2}
\begin{table}[t]
\centering
\begin{tabular}{ l   c   c  c }
 & Classification & Word Detection & Word Correction  \\ \hline 
IC - Wang~\cite{wu2017visual} & 42.21 & 27.59 & 22.16\\
HieCoAtt~\cite{lu2016hierarchical} & 64.14 & 38.79 & 4.21\\
\hline
Grounding model~\cite{hu2017modeling} & 56.68 & 39.80 & 8.80 \\
Phrase Critic (Ours) & \textbf{87.00} & \textbf{73.72} & \textbf{49.60}  \\ 
\end{tabular}
\caption{Quantitative FOIL results: Our phrase critic significantly outperforms the state-of-the-art~\cite{wu2017visual,lu2016hierarchical} reported in~\cite{shekhar2017foil} and the Grounding model~\cite{hu2017modeling} on all three FOIL tasks.}
\label{tab:foil-exps}
\end{table}
}

\begin{figure*}[t]
\centering
\includegraphics[trim={0 10 0 0},width=\linewidth]{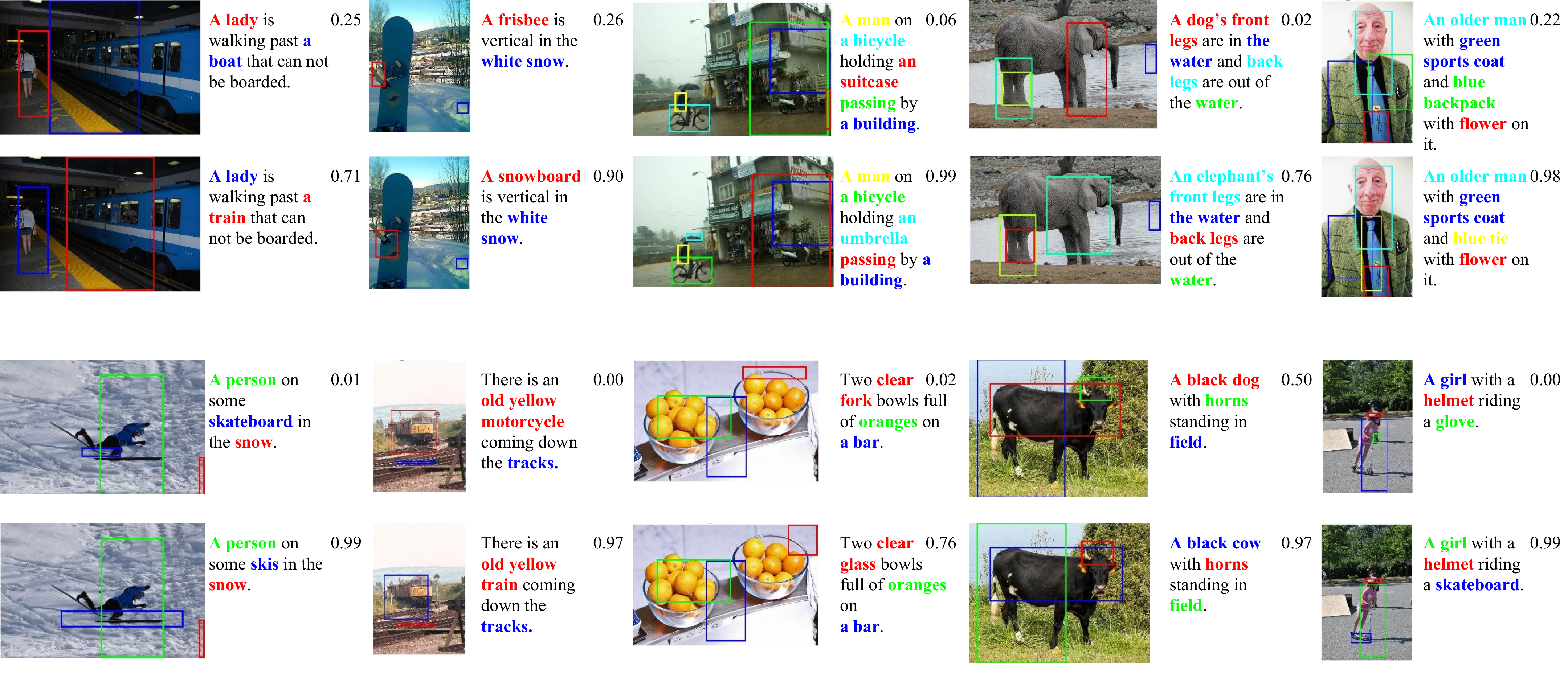}
\caption{Qualitative FOIL results: We present the image with foil sentence (top) and correct sentence (bottom) as determined by our phrase-critic model. The numbers indicate our phrase-critic score of the given sentence. By design our model grounds all the phrases in the sentence, including the foil phrases.}
\label{fig:FOIL}
\end{figure*}

\subsection{FOIL Experiments}

Table~\ref{tab:foil-exps} shows the performance of our phrase critic on the FOIL tasks compared to the best performing models evaluated in~\cite{shekhar2017foil}.
IC-Wang~\cite{wu2017visual} is an image captioning model whereas HieCoAtt~\cite{lu2016hierarchical} is an attention based VQA model.
As described above, we follow the protocol of~\cite{shekhar2017foil} when evaluating our model on the FOIL tasks.
To apply the grounding model to the classification task, we determine a threshold score on the train set (i.e., any sentence with an average grounding score above a certain threshold is classified as image relevant).

Our results show that the phrase critic is able to effectively adapt a grounding model in order to determine whether or not sentences are image relevant.
We see that our grounding model baseline performs competitively when compared to prior work, indicating that grounding noun phrases is a promising step to determine if sentences are image relevant.
However, our phrase critic model outperforms all baselines by a wide margin, outperforming the next best model by over 20 points on the classification task, over 30 points on the word identification task, and close to 30 points on the word correction task.
The large gap between the grounding model baseline and the phrase-critic highlights the importance of our phrase-critic in learning how to properly adapt outputs from a grounding model to our final task.

\autoref{fig:FOIL} shows example negative and positive sentences from the FOIL dataset, the grounding determined by our phrase-critic, and the score output by our phrase-critic model.  Our general observation is that our phrase critic gives a significantly lower score to FOIL sentences which are not image relevant. In addition, it accurately grounds mentioned objects and accurately scores sentences based on if they are image relevant. 

Some detailed observations are as follows. For the first example the score of the FOIL sentence is $0.25$ as the sentence contains ``a boat'' phrase that is inaccurate whereas the sentence with the correct phrase, i.e. ``a train'', gets the score $0.71$ which clearly indicates that this is the correct sentence. Our model is able to ground more than two phrases accurately as well. For the last image in the first row, the phrases ``an older man'', ``green sports coat'' and ``flower'' are correctly predicted and grounded whereas ``blue backpack'' gets grounded close to the shoulder, which is a sensible region to consider even though there is no backpack in the image. This FOIL sentence gets the score $0.22$ whereas the correct sentence that gets the score $0.98$ grounds ``blue tie'' correctly while also correctly grounding all other phrases in the sentence. 

When the FOIL object is one of the many objects in the sentence and occupies a small region in the image, our phrase-critic is also successful. For instance in the third image in the first row, ``an suitcase'' is grounded in an arbitrary location on the side of an image which leads to an extremely low sentence score, $0.06$. In the image relevant sentence, ``an umbrella'' is grounded correctly leading to a high sentence score, $0.99$. In conclusion, our phrase critic accurately grounds the phrases when they are present and assigns scores to the matching phrases and bounding boxes that helps us further understand why a model has taken such a decision.

\section{Conclusions}
We propose a phrase-critic model which measures the image-relevance of a generated explanation sentence.
In this framework, we first generate alternative natural language explanations of a fine-grained visual classification decision, then factorize the explanation sentences into a set of visual attributes and visually ground them. Our phrase critic model (1) assigns normalized scores to noun phrases measuring how well they are grounded in the image, (2) ranks the sentences based on the cumulative score, (3) selects the best explanation that is both image and class relevant.

Our experiments on the CUB dataset shows that this grounding approach forces our explanations to refer to the elements that are present in the image and are class-specific. We evaluate our model on the fine-grained bird explanation task both qualitatively and quantitatively. 
An intuitive alternative for factual explanations, i.e. explaining the correct class, is counterfactual explanations, i.e. explaining why a certain image is not classified as another class. We show that our model is able to generate accurate counterfactual explanations.
Our experiments on the FOIL tasks illustrates that our model is able to determine an inaccurate phrase in the sentence, point to the evidence in the image and also correct it significantly better than the competing models.

\paragraph{Acknowledgements.} This work was supported by DARPA XAI program.

\bibliographystyle{splncs04}
\bibliography{egbib}
\end{document}